\documentclass{article}
\usepackage{xcolor}

\usepackage[preprint]{corl_2026} 
\usepackage{amssymb}  
\usepackage{hyperref}
\usepackage{color,soul}
\usepackage{amsmath}
\usepackage{amsfonts}
\usepackage{hyperref}
\usepackage{graphicx}
\graphicspath{{figures-compact/}}
\usepackage{algorithm}
\usepackage{algpseudocode}
\usepackage{fancyhdr}

\title{\LARGE \bf
Agentic Aerial Cinematography:\\%
From Dialogue Cues to Cinematic Trajectories}
\author{
  \textbf{Yifan Lin}$^{\dagger *}$,
  \textbf{Sophie Z. Liu}$^{*}$,
  \textbf{Ran Qi}$^{*}$,
  \textbf{George Z. Xue}$^{*}$\\
  \textbf{Xinping Song},
  \textbf{Chao Qin},
  \textbf{Hugh H.-T. Liu}\\
  University of Toronto\\
  {\small $^{\dagger}$Project lead, $^{*}$Equal contribution (alphabetical order)}
}

\hypersetup{
  pdftitle={Agentic Aerial Cinematography: From Dialogue Cues to Cinematic Trajectories},
  pdfauthor={Yifan Lin, Sophie Z. Liu, Ran Qi, George Z. Xue, Xinping Song, Chao Qin, Hugh H.-T. Liu},
  pdfsubject={Autonomous aerial cinematography},
  pdfkeywords={Agentic Robotics, Language-Guided Planning}
}

\begin{document}
\maketitle
\suppressfloats[t]


\begin{abstract}
Agents have demonstrated strong capabilities in digital environments, yet their reasoning abilities alone are insufficient to accomplish tasks in the physical world. We present Agentic Aerial Cinematography: From Dialogue Cues to Cinematic Trajectories (ACDC), an agentic framework that bridges this gap for autonomous aerial cinematography. ACDC provides an execution harness that grounds an agent’s interpretation of cinematic intent and visual judgment in scene geometry and flight constraints. Given a natural-language request and a 3D scene representation, the agent selects relevant viewpoints and guides their refinement through visual feedback, while optimization and motion planning translate these decisions into executable drone trajectories. Experiments across a variety of simulated indoor scenes and hardware-in-the-loop flight tests demonstrate how this framework connects open-ended user intent to prompt-aligned video tours and physical drone execution. Project website: \url{https://acdc-drone.github.io}.

\end{abstract}

\keywords{Agentic Robotics, Language-Guided Planning} 


\section{Introduction}

\begin{figure}[t]
  \centering
  \includegraphics[width=\textwidth]{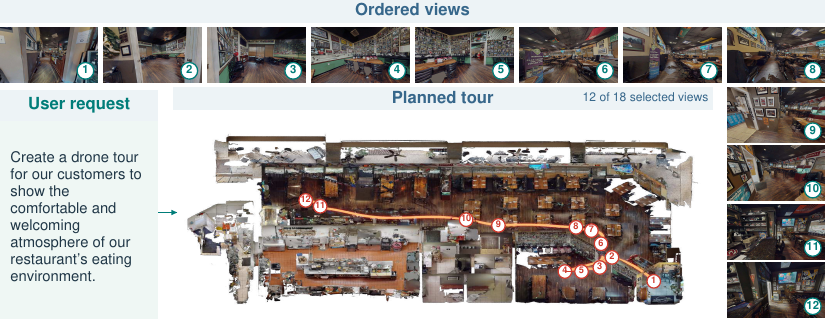}
  \caption{\textbf{The aerial cinematography task addressed by ACDC.} Given an underspecified natural-language instruction (left) and a posed exploration video of the scene, ACDC plans an ordered sequence of camera viewpoints and a quadrotor flight path through them. In this restaurant example, the TOGT-optimized trajectory (red) connects 18 selected viewpoints; 12 sampled views are shown and numbered along the top and right.}
  \label{fig:task}
\end{figure}

Unmanned aerial vehicles (UAVs) are widely used in cinematography, yet human pilots remain heavily involved and task-level autonomy is limited~\citep{Naegeli2017RAL,Bonatti2020JFR,Pueyo2024TRO}. A central challenge is that users describe the desired visual outcome without specifying the camera motion needed to achieve it. Consider the request, ``Shoot a video that shows how spacious this house feels.'' An autonomous agent must identify which parts of the scene convey spaciousness, choose viewpoints that capture them, and arrange these viewpoints into a flight that avoids obstacles. Understanding the instruction is therefore only one part of the task; the system must also translate that understanding into executable motion. We consider the autonomous drone cinematography task illustrated in Fig.~\ref{fig:task}, where a user specifies cinematic intent through open-vocabulary natural language and the system produces a video that reflects that intent.

Recent advances in Large Language Models (LLMs) and Vision Foundation Models (VFMs) enable embodied agents to interpret instructions and scene content, allowing users to express goals without explicit programming~\citep{Ahn2022SayCan,PaLME,LMNav}. These capabilities provide a basis for deciding what to film and assessing whether a view matches the request. They do not, by themselves, determine whether a camera pose is reachable or whether the motion between viewpoints respects obstacle clearance and robot dynamics~\citep{FMPlanner,LEVIOSA}. Applying such agents to indoor cinematography therefore requires mechanisms for connecting semantic decisions to scene geometry, evaluating candidate views, and producing physically executable motion. The challenge is to support the model's interpretation and visual judgment with the optimization and control components needed to carry out its decisions.

\begin{figure*}[t]
  \centering
  \includegraphics[width=\textwidth]{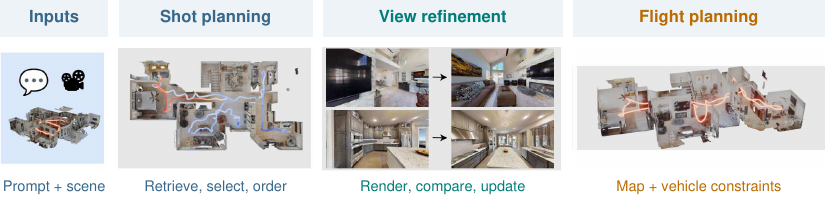}
  \caption{\textbf{System overview of ACDC.} Given a free-form prompt, a posed exploration video, and a renderable scene (\emph{Inputs}), ACDC operates in three stages: (i)~\emph{semantic shot planning and seed retrieval} (Sec.~\ref{sec:waypoint}) grounds VLM-generated shot descriptions in posed exploration frames; (ii)~\emph{visual preference refinement} (Sec.~\ref{sec:refine}) adjusts each seed's position and yaw by Bayesian optimization driven by VLM pairwise preferences between rendered views; and (iii)~\emph{physical trajectory generation} (Sec.~\ref{sec:traj}) connects the refined viewpoints under voxel-map and dynamic constraints.}
  \label{fig:agent-overview}
\end{figure*}

We present \textbf{A}gentic Aerial \textbf{C}inematography: From \textbf{D}ialogue Cues to \textbf{C}inematic Trajectories (ACDC), a framework for autonomous drone cinematography. Our key idea is to provide an \textit{execution harness} that connects model-based shot selection and visual evaluation to camera-pose optimization and constrained motion planning. The system takes a free-form prompt, a posed exploration video, and a high-fidelity 3D environment model as inputs, and proceeds in three stages (Fig.~\ref{fig:agent-overview}).

First, the agent interprets the request and uses vision--language retrieval to select and order candidate viewpoints from the exploration video~\citep{Liang2024KeyVideoLLM,QFrame}. Second, preference-based Bayesian optimization proposes alternative camera poses, while the vision--language model compares their rendered views against the requested content and aesthetic goals~\citep{Gonzalez2017PBO,Chu2005GPPreference,BradleyTerry1952Biometrika}. These comparisons guide iterative pose refinement. Third, a motion planner connects the refined viewpoints in the specified order, generating a smooth trajectory subject to collision-avoidance and dynamic constraints. An onboard controller then tracks the resulting trajectory. Through this division of responsibilities, users specify the intended video in natural language while ACDC handles viewpoint selection, visual refinement, and flight planning.

Our contributions are summarized below:

\begin{itemize}
  \item We introduce \textbf{ACDC}, an agentic framework that connects free-form cinematic intent to executable indoor UAV video tours through semantic viewpoint selection, visual feedback, and motion planning.

  \item We develop a \textbf{language-conditioned viewpoint-refinement method} based on preference-based Bayesian optimization, using model comparisons of rendered views to refine initial camera poses toward the requested aesthetic goals.

  \item We evaluate ACDC through \textbf{simulation and hardware-in-the-loop experiments}, examining prompt alignment, pose-refinement robustness, and execution of the generated trajectories on a physical quadrotor.
\end{itemize}

\section{Related Work}
\label{sec:related}

\subsection{Interfaces Between Reasoning and Physical Execution}
Embodied agents rely on interfaces that connect semantic decisions to perception, planning, and control. Code as Policies expresses language-conditioned behavior as programs that compose perception and control APIs~\citep{CodeAsPolicies2023}. VoxPoser constructs spatial value maps for motion planning~\citep{VoxPoser2023}, while PRoC3S searches for continuous skill parameters that satisfy geometric and physical constraints, returning infeasible proposals to the language model for revision~\citep{PRoC3S}. These systems establish the role of specialized execution components in grounding model decisions. ACDC adopts this principle for cinematic intent: shot descriptions and observed poses connect semantic choices to scene geometry, visual comparisons guide viewpoint optimization, and a motion planner generates flight trajectories. Within this execution harness, the agent guides continuous search by judging the appearance of candidate views.

\subsection{Visual Preferences for Continuous Optimization}
Visual goals often admit many satisfactory outcomes, making preferences useful for guiding optimization. Gaussian-process preference learning and preferential Bayesian optimization provide tools for modeling latent objectives and searching from comparisons~\citep{Chu2005GPPreference,Gonzalez2017PBO}. Brochu et al. apply preference-based optimization to procedural animation, using human judgments of visual examples to tune continuous parameters~\citep{Brochu2010SCA}. In robotics, RL-VLM-F queries a VLM for language-conditioned preferences between image observations and uses them to learn a reward function for reinforcement learning~\citep{RLVLMF2024}. ACDC uses such judgments to search over camera position and yaw in a given scene during planning. Each comparison updates a preference model that guides the next candidate evaluation, and the refined poses become goals for trajectory generation. The visual loop operates over renderings before flight; motion planning handles geometric and dynamic constraints.

\subsection{Language-Guided Cinematography}
Autonomous cinematography systems optimize framing and camera motion subject to geometric constraints~\citep{Naegeli2017RAL,Pueyo2024TRO}. Language-guided methods extend these capabilities to requests about scene content and visual appearance. PhotoBot retrieves reference images and transfers their compositions to a robot camera~\citep{PhotoBot}, while SplaTraj optimizes viewing costs over a semantic scene representation~\citep{SplaTraj}. ChatCam uses an LLM agent to compose camera trajectories, retrieving initial anchor views and refining their camera parameters through rendered image--text similarity~\citep{ChatCam}. CinemaTraj grounds an agent's choice of parametric camera movements in a 3D scene graph and optimizes the trajectories for collision and occlusion avoidance~\citep{CinemaTraj2026}. These methods already connect language to scene-grounded camera motion. ACDC refines retrieved viewpoints through pairwise visual judgments and Bayesian optimization, then connects them under quadrotor flight constraints. Compared with ChatCam's similarity objective, this refinement uses a preference model learned from comparisons conditioned on the intended shot. The contribution is an execution harness that brings semantic shot selection, comparative visual evaluation, and physical trajectory generation together for autonomous aerial cinematography.

\section{Methodology}
\label{sec:method}

ACDC takes a natural-language prompt $\mathcal{P}$, exploration images $\mathcal{V}=\{I_t\}_{t=1}^T$ with poses $\{\mathbf{T}_t\}\subset\mathrm{SE}(3)$, a voxel traversability map $\mathcal{M}$, and a renderable scene $\mathcal{S}$. All spatial inputs share a world frame. Our experiments use posed observations rendered from indoor scene assets; online reconstruction remains a direction for future work.

The execution harness connects the agent's semantic decisions and visual judgments to geometric planning through the three stages in Fig.~\ref{fig:agent-overview}. A VLM acts as both shot planner and visual critic: it describes and selects what to film, then compares candidate views. Shot descriptions and observed poses ground these decisions in the scene, while optimization, motion planning, and control determine how to realize the selected views.

\subsection{Semantic Shot Planning and Seed Retrieval}
\label{sec:waypoint}

The agent first summarizes the exploration video into an environment description $\mathcal{E}$ and, conditioned on $(\mathcal{P},\mathcal{E})$, generates $n$ shot descriptions $\{\mathcal{D}_k\}_{k=1}^n$. Each description expresses the content and visual qualities sought in one shot and serves as the goal for both retrieval and subsequent refinement. A shared vision--language encoder embeds each description as $\mathbf{d}_k=\phi_{\mathrm{txt}}(\mathcal{D}_k)$ and each frame as $\mathbf{i}_t=\phi_{\mathrm{img}}(I_t)$. For each description, cosine similarity retrieves up to eight highest-scoring frames, following language-conditioned keyframe selection~\citep{Liang2024KeyVideoLLM,QFrame}.

\begin{figure}[t]
  \centering
  \includegraphics[width=\textwidth]{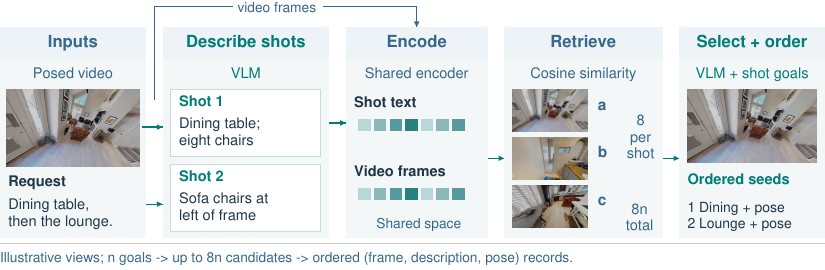}
  \caption{\textbf{Semantic shot planning and seed retrieval.} A VLM parses the exploration video and instruction into $n$ waypoint descriptions; text and frame embeddings are matched to retrieve up to eight candidate frames per description; the VLM selects and orders a subset of the candidate views, yielding (frame, description, pose) triples.}
  \label{fig:selection}
\end{figure}

The agent examines the retrieved frames, selects suitable views, and orders them according to the intended tour (Fig.~\ref{fig:selection}). Descriptions without a suitable view may be omitted; hereafter $n$ denotes the retained number of views. The resulting sequence $\{(\mathbf{T}_{k,0},\mathcal{D}_k)\}_{k=1}^n$ pairs each shot goal with an observed camera pose. Selecting an image therefore also selects a known location and viewing direction. Each refinement seed $\mathbf{x}_{k,0}=(\mathbf{p}_{k,0},\theta_{k,0})$ retains that frame's position and yaw. This grounds the initial choices in observed scene content before the optimizer searches for improved views beyond the exploration path. Appendix~\ref{app:retrieval} gives the retrieval steps and representation details.

\subsection{Visual Preference Refinement}
\label{sec:refine}

The refinement stage lets the agent guide camera placement by evaluating visual outcomes. For each shot, we search over position and yaw, $\mathbf{x}=(\mathbf{p},\theta)\in\mathbb{R}^3\times S^1$, with roll and pitch fixed to zero. A separate Gaussian process (GP) models the latent visual utility $f_k(\mathbf{x})$ under description $\mathcal{D}_k$. Its prior relates nearby positions and viewing directions; its posterior is conditioned on the comparisons collected for that shot. Pairwise preferences provide the optimization signal without a manually specified aesthetic cost~\citep{Chu2005GPPreference,Gonzalez2017PBO,Brochu2010SCA}.

At iteration $b$, the optimizer samples a set $X_b$ of $M$ poses around the incumbent $\hat{\mathbf{x}}$, within translation and yaw radii $\rho_{\mathrm{trans}}$ and $\rho_{\mathrm{rot}}$. It draws one joint sample $\tilde f$ from the current preference posterior on $\{\hat{\mathbf{x}}\}\cup X_b$ and selects the challenger with the largest sampled advantage:
\begin{equation}
\mathbf{x}^{\star}=\arg\max_{\mathbf{x}_m\in X_b}\;\tilde f(\mathbf{x}_m)-\tilde f(\hat{\mathbf{x}}).
\label{eq:challenger}
\end{equation}
This Thompson-sampling rule uses both the estimated utility and posterior uncertainty to explore candidate poses~\citep{Gonzalez2017PBO}. Only the selected challenger is submitted for comparison, requiring one VLM query per iteration.

The VLM acts as a visual critic $\Phi$, comparing the challenger and incumbent against the shot description:
\begin{equation}
s_b=\Phi\!\left(R_{\mathcal{S}}(\mathbf{x}^{\star}),R_{\mathcal{S}}(\hat{\mathbf{x}}),\mathcal{D}_k\right)\in\{+1,-1\},
\label{eq:critic}
\end{equation}
where $R_{\mathcal{S}}$ renders a pose in the scene and $s_b=+1$ indicates a preference for the challenger. A Bradley--Terry likelihood~\citep{BradleyTerry1952Biometrika} links the comparison to the latent utility:
\begin{equation}
p(s_b\mid f_k)=\sigma\!\left(s_b[f_k(\mathbf{x}^{\star})-f_k(\hat{\mathbf{x}})]\right),
\qquad \sigma(z)=(1+e^{-z})^{-1}.
\label{eq:preference}
\end{equation}

The challenger, incumbent, and preference label are added to a comparison history for the current shot. Because the likelihood is non-Gaussian, we approximate the posterior with a Gaussian and recompute it after each comparison~\citep{Chu2005GPPreference}. Histories remain separate across shot descriptions, since each defines a different visual goal. A preferred challenger replaces the incumbent. The sampling neighborhood follows the incumbent, so accepted moves can extend beyond the initial search radius. After $B$ comparisons, the incumbent is returned as $\mathbf{x}_{k,\mathrm{ref}}$, the most recent comparison winner rather than the maximizer of the posterior mean. The description remains fixed throughout refinement, linking pose changes to the same cinematic goal. Feedback operates over rendered views before flight; physical feasibility is assessed by the downstream planner. Appendix~\ref{app:alg} specifies the kernel, sampling distributions, posterior update, and full algorithm.

\subsection{Physical Trajectory Generation}
\label{sec:traj}

The ordered refined viewpoints specify the views to be visited. The motion layer determines how to connect them using the supplied map and vehicle constraints. It first searches for a path through free voxels in $\mathcal{M}$ using $\mathrm{A}^{\star}$. Rectangular corridors are grown around the path using a distance transform to retain an obstacle-clearance margin. The Time-Optimal Gate-Traversing (TOGT) planner~\citep{togt} treats these corridor boxes as gates and computes a position trajectory $\mathbf{p}(t)$ under quadrotor dynamic limits.

Because TOGT does not plan yaw, we unwrap the viewpoint yaws and interpolate them with a $C^2$ cubic spline $\theta(t)$ at their waypoint passage times. The resulting position and yaw references are passed to the tracking controller. During flight, the controller uses state feedback to follow this trajectory; the semantic shot plan and visual refinement are completed beforehand. Feasibility depends on the supplied map and vehicle model, and visually preferred poses are not automatically reachable. Appendix~\ref{app:trajectory} gives the path, corridor, and interpolation construction.

\section{Experiments}
\label{sec:exp}

We design the experiments to answer three questions, one for each stage of ACDC:
\begin{itemize}
  \item[\textbf{Q1}] Does \emph{semantic shot planning and seed retrieval} select scene content that matches prompts ranging from explicit shot lists to abstract aesthetic intent? (Sec.~\ref{sec:exp-q1})
  \item[\textbf{Q2}] Does \emph{visual preference refinement} move locally perturbed seeds toward a target view using only VLM pairwise feedback? (Sec.~\ref{sec:exp-q2})
  \item[\textbf{Q3}] Can trajectories from \emph{physical trajectory generation} be tracked by a physical quadrotor? (Sec.~\ref{sec:exp-q3})
\end{itemize}
These experiments include hardware-in-the-loop (HIL) flight to validate physical execution.

\subsection{Experimental Setup}
\label{sec:impl}

We render three indoor scenes from the HM3D dataset~\citep{ramakrishnan2021hm3d} (cottage, restaurant, luxury) in Unity. Exploration videos are recorded by a human operator navigating each rendered scene with a motion-captured handheld display, and every frame is stored with its camera pose; retrieval, rendering, and trajectory planning share the same scene frame (Appendix~\ref{app:impl}).

In both semantic shot planning (Sec.~\ref{sec:waypoint}) and visual preference refinement (Sec.~\ref{sec:refine}), we use GPT-6~Astra as the vision–language model and AM-RADIO~\citep{Ranzinger_2024_CVPR} with a SigLIP~2 adaptor~\citep{tschannen2025siglip2multilingualvisionlanguage} as the pretrained visual encoder. During refinement, each waypoint is optimized for $B=100$ iterations (one VLM comparison per iteration) with $M=64$ candidates per iteration and sampling radii $\rho_{\mathrm{trans}}=1\,\mathrm{m}$ and $\rho_{\mathrm{rot}}=30^\circ$.
The HIL platform tracks the planned position and yaw references with a nonlinear model predictive contouring controller (MPCC).

\begin{figure*}[t]
  \centering
  \includegraphics[width=\textwidth]{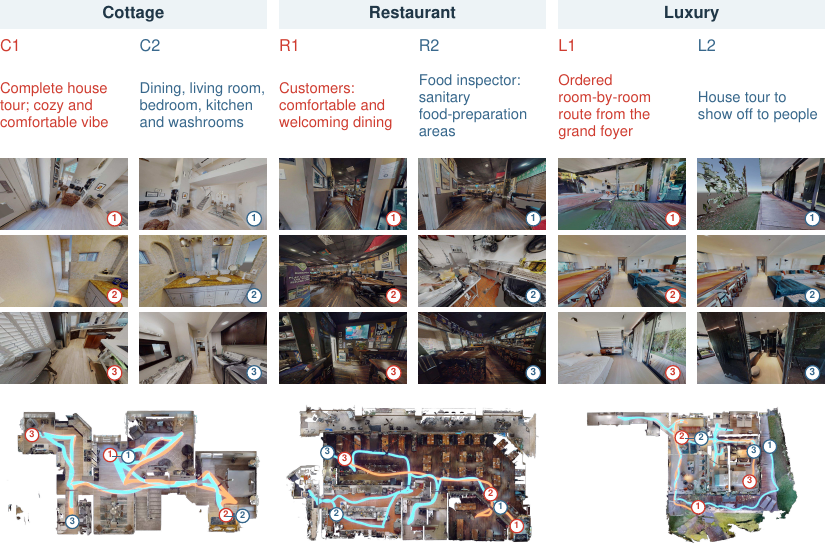}
  \caption{\textbf{Qualitative ACDC tours in three simulated scenes (Q1).} Columns correspond to the \emph{cottage} (left), \emph{restaurant} (middle), and \emph{luxury} (right) scenes. For each scene, two tours are shown with their prompt summaries and three representative keyframes per tour. Both TOGT-optimized position trajectories are overlaid on each scene at the bottom; thumbnails show the original selected exploration frames.}
  \label{fig:main_result}
\end{figure*}

\begin{figure}[t]
  \centering
  \begin{minipage}[t]{0.48\textwidth}
    \vspace{0pt}\centering
    \parbox[c][2.45in][c]{\linewidth}{\centering\includegraphics[width=\linewidth,height=2.45in,keepaspectratio]{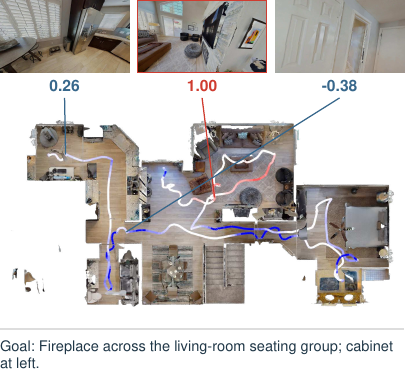}}
    \caption{\textbf{Retrieval heatmap (Q1).} Exploration path colored by the cosine similarity between each frame embedding and the shot description below (red: high; blue: low). Colors use similarity divided by the highest similarity, clipped to $[0,1]$. Insets show three frames with their relative similarity scores (highest $=1.00$); the highest-scoring frame (red border) shows the described fireplace and living-room seating group.}
    \label{fig:cosine}
  \end{minipage}\hfill
  \begin{minipage}[t]{0.48\textwidth}
    \vspace{0pt}\centering
    \parbox[c][2.45in][c]{\linewidth}{\centering\includegraphics[width=\linewidth,height=2.45in,keepaspectratio]{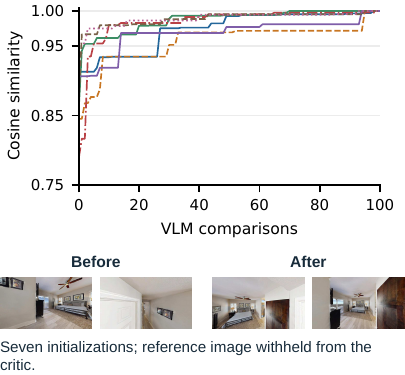}}
    \caption{\textbf{Visual preference refinement from perturbed seeds (Q2).} Image-embedding cosine similarity between the incumbent view and the reference view at $\mathbf{T}_{\mathrm{tgt}}$ over 100 iterations for seven initializations within $1\,\mathrm{m}$ of $\mathbf{T}_{\mathrm{tgt}}$. All traces approach $\approx 1.0$, i.e., the refined views approach the reference view in embedding space. Thumbnails: views before and after refinement.}
    \label{fig:refinement_robustness}
  \end{minipage}
\end{figure}

\subsection{Q1: Semantic Shot Planning and Seed Retrieval}
\label{sec:exp-q1}

Figure~\ref{fig:main_result} shows six tours across cottage, restaurant, and luxury scenes. The prompts include explicit shot instructions, situation-dependent requests, and abstract aesthetic intent. The cottage example follows an ordered request; the restaurant examples select different areas for a customer and a food inspector; and the luxury examples pair a guest walkthrough with an abstract request to highlight the modern design. These cases show how semantic interpretation changes the selected content and framing, but are qualitative examples rather than a systematic test of cross-scene generalization.

To inspect retrieval, we compute cosine similarity between each exploration-frame embedding and the selected shot-description embedding, then project the scores onto the floor plan (Fig.~\ref{fig:cosine}). High-scoring regions coincide with the requested area in the illustrated example, and the insets show retrieved frames. This visualizes the semantic grounding step.

\subsection{Q2: Visual Preference Refinement}
\label{sec:exp-q2}

We select a reference pose $\mathbf{T}_{\mathrm{tgt}}$ and initialize seven refinement runs at positions within $1\,\mathrm{m}$ of it, with randomized yaw. Each run uses the description $\mathcal{D}_{\mathrm{tgt}}$ generated for the reference view and a budget of $B=100$ iterations. At each iteration $b$, we embed the rendered incumbent view $R_{\mathcal{S}}(\hat{\mathbf{x}})$ and the reference image $I_{\mathrm{tgt}}$ using $\phi_{\mathrm{img}}$ and measure their cosine similarity. The critic $\Phi$ never sees $I_{\mathrm{tgt}}$; it compares candidates only against $\mathcal{D}_{\mathrm{tgt}}$. All seven reported traces approach a similarity of approximately one within the budget (Fig.~\ref{fig:refinement_robustness}), including initial views in which the target is outside the image.

These results show that description-conditioned pairwise feedback can guide locally perturbed viewpoints toward the reference view in embedding space, supporting refinement as the link between semantic shot goals and continuous camera-pose search. The evidence is limited to the tested initializations and reference-view similarity metric; it does not establish exact pose recovery or recovery from physical execution failures.

\subsection{Q3: Hardware-in-the-Loop Execution}
\label{sec:exp-q3}

We conduct hardware-in-the-loop tests using the quadrotor in Fig.~\ref{fig:hitl}, equipped with an NVIDIA Jetson Orin NX running the autopilot stack. Motion-capture pose estimates are fused with onboard IMU measurements for six-degree-of-freedom odometry. The illustrated flight demonstrates execution of an ACDC trajectory on the physical platform. It supports the trajectory-to-controller interface, while the scene and cinematic views remain part of the simulated-scene evaluation.

\section{Discussion and Limitations}
\label{sec:discussion}

We observed a recurring loss of task focus in generated videos: the camera leaves relevant content and spends part of the sequence showing areas with little connection to the request. Individual shots can appear reasonable while their arrangement produces a fragmented account of the scene. We suspect this reflects a limitation in how spatial reasoning is used across the sequence: choosing locally suitable views does not by itself connect where requested content is located, which areas have been covered, and where the camera should go next.

\begin{figure}[t]
  \centering
  \begin{minipage}[t]{0.48\textwidth}
    \vspace{0pt}\centering
    \parbox[c][1.82in][c]{\linewidth}{\centering\includegraphics[width=\linewidth]{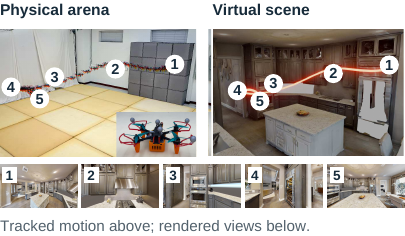}}
    \caption{\textbf{Hardware-in-the-loop flight (Q3).} Tracked flight in the motion-capture arena (left), its simulated rendering (right), and waypoint views (bottom).}
    \label{fig:hitl}
  \end{minipage}\hfill
  \begin{minipage}[t]{0.48\textwidth}
    \vspace{0pt}\centering
    \parbox[c][1.82in][c]{\linewidth}{\centering\includegraphics[width=\linewidth]{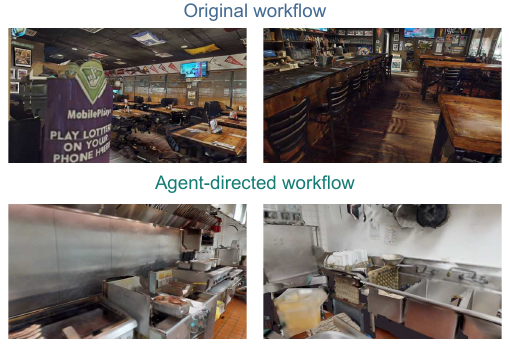}}
    \caption{\textbf{Key frames with GPT-6 Astra.} Prompt: ``Create a virtual drone tour for a food safety inspector, showing that our restaurant maintains a highly sanitary food processing environment.'' Original workflow (top) includes off-topic views; agent-directed workflow (bottom) stays on task. Both use Astra; frames show key moments from the generated tour videos.}
    \label{fig:agent-finding}
  \end{minipage}
\end{figure}

\textbf{Greater workflow freedom with GPT-6 Astra.} With recent VLMs offering strong spatial understanding, we gave the agent greater freedom to use this capability across the planning workflow. We revisited the same scene-prompt tasks with the recently released GPT-6 Astra~\citep{OpenAI2026AstraRelease}, allowing it to construct and revise its own workflow around the request. The agent could choose how to use scene information and planning tools, inspect rendered outputs, and reorganize coverage, routes, camera direction, and timing together. The resulting sequence stays with the requested content and removes the off-topic detours visible in the original output (Fig.~\ref{fig:agent-finding}). This qualitative improvement illustrates the value of treating the video as a connected spatial sequence: scene layout can guide both the selection of relevant views and how the camera moves between them.

The framework assumes a static, mapped scene and depends on exploration coverage, rendering quality, and model judgments. Controlled human comparisons and physical flight tests remain important next steps for assessing cinematic quality and execution.

\section{Conclusion}

We presented ACDC, an execution harness for autonomous aerial cinematography that connects semantic shot selection and visual preference refinement to constrained drone flight. Simulated tours illustrate alignment with varied cinematic requests, while local refinement experiments show recovery toward reference views in embedding space. Hardware-in-the-loop flight demonstrates execution of the planned trajectories on a physical quadrotor. ACDC thus provides a concrete example of how an agent's interpretation and visual judgment can be supported by optimization, motion planning, and control to carry out open-ended cinematic goals.


\acknowledgments{The authors thank Wenda Zhao for experimental support and H S Helson Go for his contribution in the development of the autopilot stack.}


\bibliography{cit_old,cit_new,cit_interfaces}  

\clearpage
\appendix
\section{Semantic Shot Planning and Retrieval Details}
\label{app:retrieval}

This section expands the retrieval procedure in Sec.~\ref{sec:waypoint}. The exploration frames provide a finite set of visual observations with known poses, grounding the initial shot selection before novel-view refinement. The video provides scene coverage rather than deliberately composed shots; using observed frames avoids relying on potentially incomplete novel-view renderings for initial grounding.

\paragraph{Waypoint Description Generation.} Given the exploration video, a VLM summarizes it into an environment description $\mathcal{E}$ and, conditioned on $(\mathcal{P}, \mathcal{E})$, generates $n$ waypoint descriptions $\{\mathcal{D}_k\}_{k=1}^n$.
\paragraph{Input Encoding.} A pretrained vision foundation model (VFM) encodes each description as $\mathbf{d}_k=\phi_{\mathrm{txt}}(\mathcal{D}_k)$ and each frame $I_t$ as $\mathbf{i}_t=\phi_{\mathrm{img}}(I_t)$ using the same model.
\paragraph{Candidate Retrieval.} For each description embedding $\mathbf{d}_k$, we compute the cosine similarity $\mathbf{d}_k^\top\mathbf{i}_t/(\|\mathbf{d}_k\|\,\|\mathbf{i}_t\|)$ with every frame embedding $\mathbf{i}_t$ and retain up to eight highest-scoring frames as the candidate set $C_k\subset\{1,\dots,T\}$ (frame indices, each carrying its pose $\mathbf{T}_t$).
\paragraph{Select and Sort.} A multimodal VLM receives the descriptions $\{\mathcal{D}_k\}_{k=1}^n$ and candidate frames $\{C_k\}_{k=1}^n$, selects suitable frames (omitting descriptions without a suitable candidate), and orders the selected views according to the intended tour.

The output is an ordered sequence of pose-description pairs $\{(\mathbf{T}_{k,0},\mathcal{D}_k)\}_{k=1}^n$, indexed after sorting. Each $\mathbf{T}_{k,0}$ is the recorded pose of its selected frame; its position and yaw define the refinement seed $\mathbf{x}_{k,0}$ (Sec.~\ref{sec:refine}), and its roll and pitch are discarded. These pairs connect semantic shot goals to spatially grounded seeds for refinement; the VLM is not asked to infer metric coordinates from text alone.

\section{Visual Preference Refinement Details}
\label{app:alg}

This section gives the preference model, candidate sampling rule, and posterior update used in Sec.~\ref{sec:refine}.

For the evaluated aerial setting, we optimize four degrees of freedom: translation $\mathbf{p}\in\mathbb{R}^3$ and yaw $\theta\in S^1$. Camera roll and pitch are fixed to zero in the viewpoint search, and orientation is represented by the yaw quaternion $\mathbf{q}=\mathbf{q}_z(\theta)$. We write a pose as $\mathbf{x}=(\mathbf{p},\theta)\in\mathcal{X}=\mathbb{R}^3\times S^1$ and identify it with $\mathbf{T}\in\mathrm{SE}(3)$ having position $\mathbf{p}$ and rotation $\mathbf{q}_z(\theta)$. Algorithm~\ref{alg:pose-refine} summarizes the preference loop.

\paragraph{GP Prior.}
For shot $k$, let $f_k:\mathcal{X}\to\mathbb{R}$ denote the latent visual utility of the view rendered at $\mathbf{x}$ with respect to $\mathcal{D}_k$. We place a zero-mean Gaussian process (GP) prior $f_k\sim\mathcal{GP}(0,\kappa)$ with a product covariance over position and yaw,
\[
\kappa(\mathbf{x}_i,\mathbf{x}_j)=\kappa_{\mathrm{rot}}(\theta_i,\theta_j)\,\kappa_{\mathrm{trans}}(\mathbf{p}_i,\mathbf{p}_j)+\varepsilon\,\delta_{ij},
\]
where $\kappa_{\mathrm{trans}}$ is a squared-exponential kernel on $\|\mathbf{p}_i-\mathbf{p}_j\|$, $\kappa_{\mathrm{rot}}$ is a stationary kernel on the wrapped angular distance $d_{S^1}(\theta_i,\theta_j)=|\mathrm{wrap}(\theta_i-\theta_j)|$ (the rotation geodesic between the yaw quaternions $\mathbf{q}_z(\theta_i)$ and $\mathbf{q}_z(\theta_j)$), and $\varepsilon\,\delta_{ij}$ is a small diagonal jitter. A new GP is used for each shot, since preferences elicited under different descriptions observe different utilities.

\paragraph{Candidate Sampling.} At each iteration $b$, we draw $M$ candidates around the incumbent $\hat{\mathbf{x}}=(\hat{\mathbf{p}},\hat\theta)$. For $m=1,\dots,M$,
\begin{equation}
\begin{aligned}
&\theta_m=\mathrm{wrap}(\hat\theta+\Delta\theta_m), && \Delta\theta_m \sim \mathrm{Unif}[-\rho_{\mathrm{rot}},\rho_{\mathrm{rot}}],\\
&\mathbf{p}_m=\mathrm{clamp}_z(\hat{\mathbf{p}}+r_m\mathbf{u}_m), && \mathbf{u}_m \sim \mathrm{Unif}(S^2),\; r_m \sim \mathrm{Unif}[0,\rho_{\mathrm{trans}}],
\end{aligned}
\label{eq:sample}
\end{equation}
and $X_b=\{\mathbf{x}_m=(\mathbf{p}_m,\theta_m)\}_{m=1}^M$. Here $\mathrm{wrap}(\cdot)$ maps angles to $(-\pi,\pi]$ and $\mathrm{clamp}_z$ clips the height to $[z_{\min},z_{\max}]$. Sampling the radius uniformly concentrates candidates near the incumbent. Because the window is centered on the current incumbent, accepted moves can accumulate beyond $\rho_{\mathrm{trans}}$ and $\rho_{\mathrm{rot}}$ from the seed.

\paragraph{Challenger Selection.} We choose the challenger by Thompson sampling against the incumbent, in the spirit of dueling Thompson sampling~\citep{Gonzalez2017PBO}: one joint sample $\tilde f$ is drawn from the current (approximate) posterior of $f_k$ on $\{\hat{\mathbf{x}}\}\cup X_b$, and
\[
\mathbf{x}^\star=\arg\max_{\mathbf{x}_m\in X_b}\; \tilde f(\mathbf{x}_m)-\tilde f(\hat{\mathbf{x}}).
\]
The sampled advantage over the incumbent balances exploitation (high posterior mean) and exploration (high posterior variance), and only one VLM query is issued per iteration.
\paragraph{Preference Feedback.}
Let $R_{\mathcal{S}}(\mathbf{x})$ denote the image rendered from $\mathcal{S}$ at pose $\mathbf{x}$. We present $R_{\mathcal{S}}(\mathbf{x}^\star)$, $R_{\mathcal{S}}(\hat{\mathbf{x}})$, and $\mathcal{D}_k$ to the VLM critic $\Phi$, which returns $s_b=+1$ if it prefers the challenger and $s_b=-1$ otherwise; the triple $(\mathbf{x}^\star,\hat{\mathbf{x}},s_b)$ is appended to the comparison history $\mathcal{H}_k$. We model each comparison with the Bradley-Terry (logistic) likelihood~\citep{BradleyTerry1952Biometrika}
\[
p\big(s_b\mid f_k\big)=\sigma\!\big(s_b\,[f_k(\mathbf{x}^\star)-f_k(\hat{\mathbf{x}})]\big),\qquad \sigma(z)=1/(1+e^{-z}),
\]
which links the probability of preferring one view to the latent utility difference.

\paragraph{Posterior Update and Output.}
Because the likelihood is non-Gaussian, the posterior $p(f_k\mid\mathcal{H}_k)$ is approximated with a Gaussian~\citep{Chu2005GPPreference} and recomputed after each comparison. If the VLM prefers the challenger ($s_b=+1$), it becomes the incumbent. After $B$ iterations, the incumbent is returned as $\mathbf{x}_{k,\mathrm{ref}}=(\mathbf{p}_{k,\mathrm{ref}},\theta_{k,\mathrm{ref}})$, giving the ordered refined poses $\{\mathbf{x}_{k,\mathrm{ref}}\}_{k=1}^n$. The incumbent is thus the most recent winner of a direct comparison, not the maximizer of the posterior mean.
This process seeks views favored by the visual critic; it does not guarantee monotonic improvement in human judgments or convergence to a global aesthetic optimum.

\begin{algorithm}[H]
\caption{Pose refinement via preference-based Bayesian optimization over $\mathbf{x}=(\mathbf{p},\theta)\in\mathbb{R}^3\times S^1$}
\label{alg:pose-refine}
\begin{algorithmic}[1]
\Require Seeds and descriptions $\{(\mathbf{x}_{k,0},\mathcal{D}_k)\}_{k=1}^n$; scene $\mathcal{S}$; VLM critic $\Phi$; iterations $B$; candidates per iteration $M$; sampling radii $(\rho_{\mathrm{rot}},\rho_{\mathrm{trans}})$
\For{$k=1$ to $n$}
  \State Initialize $f_k \sim \mathcal{GP}(0,\kappa)$ on $\mathcal{X}$; $\mathcal{H}_k\gets\emptyset$ \Comment{one utility per shot}
  \State $\hat{\mathbf{x}} \gets \mathbf{x}_{k,0}$ \Comment{incumbent}
  \For{$b=1$ to $B$}
    \State $X_b \gets \textsc{SampleCandidates}(\hat{\mathbf{x}}, M, \rho_{\mathrm{rot}}, \rho_{\mathrm{trans}})$ \Comment{Eq.~\eqref{eq:sample}}
    \State $\tilde f \sim p(f_k \mid \mathcal{H}_k)$ evaluated on $\{\hat{\mathbf{x}}\}\cup X_b$ \Comment{Thompson draw}
    \State $\mathbf{x}^\star \gets \arg\max_{\mathbf{x}\in X_b}\, \tilde f(\mathbf{x})-\tilde f(\hat{\mathbf{x}})$ \Comment{challenger}
    \State $s_b \gets \Phi\big(R_{\mathcal{S}}(\mathbf{x}^\star), R_{\mathcal{S}}(\hat{\mathbf{x}}), \mathcal{D}_k\big)\in\{+1,-1\}$ \Comment{VLM comparison}
    \State $\mathcal{H}_k \gets \mathcal{H}_k \cup \{(\mathbf{x}^\star,\hat{\mathbf{x}},s_b)\}$; update $p(f_k\mid\mathcal{H}_k)$
    \If{$s_b=+1$}
      \State $\hat{\mathbf{x}} \gets \mathbf{x}^\star$
    \EndIf
  \EndFor
  \State $\mathbf{x}_{k,\mathrm{ref}} \gets \hat{\mathbf{x}}$
\EndFor
\State \Return $\{\mathbf{x}_{k,\mathrm{ref}}\}_{k=1}^n$
\end{algorithmic}
\end{algorithm}

\section{Physical Trajectory Generation Details}
\label{app:trajectory}

This section gives the geometric construction and yaw interpolation summarized in Sec.~\ref{sec:traj}.

The motion layer takes refined positions $\mathbf{p}_{k,\mathrm{ref}}$, yaws $\theta_{k,\mathrm{ref}}$, and the traversability map $\mathcal{M}$. It connects the viewpoints using geometric planning and quadrotor constraints, independently of the VLM's aesthetic judgments. Feasibility is assessed relative to the supplied map and model; preferred viewpoints are not automatically reachable.

\paragraph{Path Extraction.}
We connect the waypoint positions in order using $\mathrm{A}^\star$ on the free-voxel graph of $\mathcal{M}$, yielding a polyline through free voxels
\[
\Pi=\{\mathbf{c}_\ell\}_{\ell=1}^{L},\qquad
\mathbf{c}_{\ell_k}=\mathbf{p}_{k,\mathrm{ref}},\quad 1=\ell_1<\ell_2<\dots<\ell_n=L,
\]
that visits every $\mathbf{p}_{k,\mathrm{ref}}$ in order.

\paragraph{Rectangular-Tunnel Corridor.}
For each segment $(\mathbf{c}_\ell,\mathbf{c}_{\ell+1})$, let $\mathbf{t}_\ell$ be the unit tangent, choose orthonormal $(\mathbf{n}_\ell,\mathbf{b}_\ell)$ spanning the orthogonal plane, and set $\Delta_\ell=\|\mathbf{c}_{\ell+1}-\mathbf{c}_\ell\|$. We grow a rectangle centered on the segment (leaving margin $\delta_{\text{safe}}$ via a distance transform) to obtain cross-section $(w_\ell,h_\ell)$. The box $\mathcal{B}_\ell$ is the set of points $\mathbf{y}$ whose local coordinates $(\xi_n,\xi_b,\xi_t)$ relative to the segment midpoint $\bar{\mathbf{c}}_\ell=\tfrac12(\mathbf{c}_\ell+\mathbf{c}_{\ell+1})$, i.e., $\mathbf{y}=\bar{\mathbf{c}}_\ell+\xi_n\mathbf{n}_\ell+\xi_b\mathbf{b}_\ell+\xi_t\mathbf{t}_\ell$, satisfy
\[
|\xi_n|\le \tfrac{w_\ell}{2},\quad |\xi_b|\le \tfrac{h_\ell}{2},\quad |\xi_t|\le \tfrac{\Delta_\ell}{2}.
\]
The corridor is the union $\mathcal{K}=\bigcup_{\ell=1}^{L-1}\mathcal{B}_\ell$. Consecutive boxes share the vertex $\mathbf{c}_{\ell+1}$, so $\mathcal{K}$ is connected, and a path contained in $\mathcal{K}$ avoids mapped obstacles with clearance $\delta_{\mathrm{safe}}$.

\paragraph{Time-Optimal Trajectory Planning.}
We treat each box $\mathcal{B}_\ell$ as a gate and use the Time-Optimal Gate-Traversing (TOGT) planner~\citep{togt} to compute a position trajectory $\mathbf{p}(t)$ that traverses the gates in order under quadrotor dynamic limits. Let $\tau_k$ denote the time at which $\mathbf{p}(t)$ passes $\mathbf{p}_{k,\mathrm{ref}}$. Because TOGT does not plan yaw, we unwrap $\{\theta_{k,\mathrm{ref}}\}$ and interpolate $\{(\tau_k,\theta_{k,\mathrm{ref}})\}_{k=1}^n$ with a $C^2$ cubic spline $\theta(t)$. The pair $(\mathbf{p}(t),\theta(t))$ is the reference passed to the tracking controller.

\section{Additional Implementation Details}
\label{app:impl}

\paragraph{Exploration data collection.} During simulated exploration, a motion-capture system tracked a handheld device displaying the rendered camera view; joystick commands supplied horizontal translation, while motion capture supplied height and orientation. We recorded each rendered frame with its camera pose. Retrieval, rendering, and trajectory planning used the same scene-coordinate frame. The voxel traversability map is an input to the planner. The motion-capture volume is $6\times6\times2.5\,\mathrm{m}$; joystick navigation permits exploration beyond this volume while retaining camera poses for each frame.

\paragraph{Qualitative workflow follow-up.}
The qualitative example in Fig.~\ref{fig:agent-finding} uses saved original and agent-directed outputs for the same inspection request. Both use GPT-6 Astra. The original route uses retrieved recorded poses without pose refinement; the agent-directed run can reuse, modify, or replace planning components and review rendered outputs. Its video uses resource-adapted EEVEE rendering. The selected figure frames come from different moments in each video and are not temporally aligned. Source videos, frame timestamps, and provenance are preserved with the figure assets.

\end{document}